\documentclass[sn-mathphys,Numbered]{sn-jnl}

\usepackage{graphicx}%
\usepackage{multirow}%
\usepackage{amsmath,amssymb,amsfonts}%
\usepackage{amsthm}%
\usepackage{mathrsfs}%
\usepackage[title]{appendix}%
\usepackage{xcolor}%
\usepackage{textcomp}%
\usepackage{manyfoot}%
\usepackage{booktabs}%
\usepackage{algorithm}%
\usepackage{algorithmicx}%
\usepackage{algpseudocode}%
\usepackage{listings}%
\usepackage{subfigure} %
\usepackage{hyperref}
\usepackage{bm}
\usepackage{threeparttable}

\theoremstyle{thmstyleone}%
\theoremstyle{thmstyletwo}%

\theoremstyle{thmstylethree}%

\begin{document}

\title[POAR: Policy Optimization with Online Abstract State Representation Learning]{POAR: Efficient Policy Optimization via Online Abstract State Representation Learning}


\author[1]{\fnm{Zhaorun} \sur{Chen}}

\author[2]{\fnm{Siqi} \sur{Fan}}

\author[2]{\fnm{Yuan} \sur{Tan}}

\author*[3]{\fnm{Liang} \sur{Gong}}\email{gongliang\_mi@sjtu.edu.cn}

\author[3]{\fnm{Binhao} \sur{Chen}}

\author[3]{\fnm{Te} \sur{Sun}}
\author[4]{\fnm{David} \sur{Filliat}}

\author[4]{\fnm{Natalia} \sur{Díaz-Rodríguez}}

\author[3]{\fnm{Chengliang} \sur{Liu}}

\affil*[1]{\orgdiv{Elmore Family School of Electrical and Computer Engineering}, \orgname{Purdue University}, \orgaddress{ \city{West Lafayette}, \city{Indiana}, \country{USA}}}

\affil[2]{\orgdiv{School of Information Science and Engineering}, \orgname{Lanzhou University}, \orgaddress{\city{Lan Zhou}, \country{China}}}

\affil[3]{\orgdiv{School of Mechanical Engineering}, \orgname{Shanghai Jiaotong University}, \orgaddress{\city{Shanghai}, \country{China}}}

\affil[4]{\orgdiv{Flowers Team}, \orgname{ENSTA Paris, Institute Polytechnique de Paris \& INRIA, \orgaddress{\city{Paris}, \country{France}}}}


\abstract{While the rapid progress of deep learning fuels end-to-end reinforcement learning (RL), direct application, especially in high-dimensional space like robotic scenarios still suffers from low sample efficiency. Therefore \emph{State Representation Learning} (SRL) is proposed to specifically learn to encode task-relevant features from complex sensory data into low-dimensional states. However, the pervasive implementation of SRL is usually conducted by a decoupling strategy in which the observation-state mapping is learned separately, which is prone to over-fit. To handle such problem, we summarize the state-of-the-art (SOTA) SRL sub-tasks in previous works and present a new algorithm called \emph{Policy Optimization via Abstract Representation} which integrates SRL into the policy optimization phase. Firstly, We engage RL loss to assist in updating SRL model so that the states can evolve to meet the demand of RL and maintain a good physical interpretation. Secondly, we introduce a dynamic loss weighting mechanism so that both models can efficiently adapt to each other. Thirdly, we introduce a new SRL prior called \emph{domain resemblance} to leverage expert demonstration to improve SRL interpretations. Finally, we provide a real-time access of state graph to monitor the course of learning. Experiments indicate that POAR significantly outperforms SOTA RL algorithms and decoupling SRL strategies in terms of sample efficiency and final rewards. We emperically verify POAR to efficiently handle tasks in high dimensions and facilitate training real-life robots directly from scratch.}




\maketitle
\section{Introduction}

While learning for robotics control has benefited from various recent advances in reinforcement learning (RL) \cite{schulman2017proximal, kaiser2019model}, learning from demonstration (LfD) \cite{finn2016guided, chen2021efficiently} and safe control \cite{garcia2015comprehensive, chen2023progressive}, the curse of high dimensionality \cite{bellman1966dynamic} and embodiment transferrability\cite{huang2020one} are the major two challenges for broader robotics generalizations and robust applications. State representation learning (SRL) \cite{bengio2013representation} has proven to be an effective approach to handle them. Contrastive methods \cite{chen2020simple, he2020momentum, bachman2019learning} learn a lower-dimensional representation by training a model to distinguish between positive and negative examples. Predictive methods such as \cite{schwarzer2020data} learn latent-space transition model by explicitly predicting the future states; \cite{raffin2019decoupling} map the high-dimensional raw observations into a compact space via reconstruction. Another line of works \cite{zakka2022xirl} incorporate representation learning objectives to address the embodiment gap. \cite{liu2018imitation, yu2018one, ding2023embodied} learns transferable representations across embodiments and \cite{martin2019variable, shao2020unigrasp} searches for shared action representations.

Most previous SRL implementations\cite{raffin2019decoupling, schwarzer2021pretraining, grill2020bootstrap, jonschkowski2014state} follow a decoupling strategy that consists of two stages: \textbf{learning state representations}, which builds a representative state space for the robotic environment from raw images, and then \textbf{learning a task-specific policy}. Extensive studies \cite{raffin2019decoupling, schwarzer2020data} show that training via SRL exhibits better performance than direct end-to-end training for some RL algorithms, in terms of sample efficiency and final convergence.  
One paradigm to SRL is to incorporates optimizing objective functions with some robotic priors (e.g. reconstruction model that consists of an auto-encoder \cite{lange2012autonomous} and the environmental model where both state representation and the transition function is learned altogether \cite{jetchev2013learning}). In \cite{jonschkowski2014state}, the authors introduced some intrinsic priors specifically for robotic scenes. However, these approaches all separately feed samples to train the SRL model before RL. Since the SRL model does not update with the exploration collected by RL agent, the learned state estimator is prone to overfit. Besides, the actual performance of SRL can be only confirmed by a posterior RL training.\par

In this study, we propose Policy Optimization via Abstract Representation (POAR), which summarizes the SRL sub-learning tasks in previous works (e.g. forward \cite{gelada2019deepmdp}, inverse \cite{guo2020bootstrap}, auto-reconstruction \cite{finn2015learning}, reward \cite{lesort2018state} prediction tasks) and integrates SRL into the RL policy phase(e.g. proximal policy optimization (PPO) \cite{schulman2017proximal}) for online adaptive policy optimization. We design the algorithm to update SRL and RL models simultaneously, where the learned representation can evolve to meet the demand of reinforcement learning and maintain a good physical interpretation. Moreover, we can retrieve the learned abstract representation in real-time to monitor the course of training and diagnose the policy. Prior works often encounter difficulty in optimizing RL and SRL goals together, since they are sensitive to hyperparameters (e.g. weight parameter of losses) and often lead either one to collapse\cite{marler2004survey}. In this work, we leverage a dynamic loss weighting strategy to guarantee convergence for online optimization of both goals.\par 

In addition, to better accustom RL algorithms to 3D robotic control scenarios, we introduce a new SRL prior which leverages expert demonstration\cite{finn2016guided, sun2021robotdrlsim}. Internalizing demonstration to diminish searching space can efficiently tackle the cold-start problem in which explorations are lengthy and expensive, especially during early training stages \cite{argall2009survey}. To do so, POAR employs a demonstration-based prior to optimize the maximum mean discrepancy (MMD) \cite{gretton2012kernel} loss to reduce the distribution similarity between learned states and expert 
trajectories, so that the represented states can capture expert's behavioral modes.\par

Experimental assessment on 3 simulation environments demonstrate that under generalized conditions, POAR exhibits greater convergence rate and better final performance than PPO baselines\cite{schulman2017proximal} and decoupling strategies\cite{raffin2019decoupling}. Our main contributions can summarize as:
\begin{enumerate}
    \item We summarize the SRL sub-learning tasks in previous SOTA works and propose a generic RL framework for high-dimensional input, which integrates SRL and RL where both models can evolve together to adapt to each other to learn a better state representation and policy.
    \item By combining external demonstration, the learned states distribution adapt to resemble the expert trajectory, which mitigates the cold-start dilemma.
    \item We provide a real-time access to learned states to better monitor the SRL model and diagnose the current policy, which facilitates interpretablity in deep RL.
    \item Through increasing sample efficiency via various approaches, POAR makes it possible to directly deploy the algorithm and train on real robots in highly stochastic and interaction-expensive environments.
\end{enumerate}

\section{Preliminaries and Problem Statements}

\subsection{Reinforcement Learning (RL)}
In RL terminology, an agent interacts with the target environment to maximize the accumulative rewards received from the environment. At each time step $t$, the agent selects an action $a_t$ corresponding to the input representation of the environment's state $s_t$. As a consequence of its action, the agent receives a scalar reward $r_{t+1}$ and transits to the next state $s_{t+1}$. This process can be reformulated as an infinite-horizon Markov decision process (MDP), defined by the tuple $(\mathcal{S}, \mathcal{A}, P, r)$, where $\mathcal{S}$ is a finite set of states, $\mathcal{A}$ is a finite set of actions, and $P : \mathcal{S}\times\mathcal{A} \times \mathcal{S} \rightarrow [0,1]$ is the transition probability distribution, $r: \mathcal{S} \rightarrow \mathbb{R}$ is the reward function. Let $\pi$ denote a stochastic policy $\pi : \mathcal{S} \times \mathcal{A} \rightarrow [0,1]$ that determines distribution of agent's action at each given state. The goal is to find an optimal control policy $\pi^*$ that maximizes the expected discounted reward $J(\pi) = \mathbb{E}_{s_0, a_0, \dots}\big[\sum_{t=0}^{T} \gamma^t r(s_t)\big]$, where $\gamma\in[0,1]$ denotes the discount factor.


\subsection{State Representation Learning (SRL)}
SRL learns a function:$f: \mathcal{O} \rightarrow \mathcal{S}$ that maps observations in high-dimension to informative low-dimensional states. While deep RL algorithms have shown that it is possible to learn a feasible control policy from raw observations, SRL can take advantage of low-dimensional and informative representations, instead of raw images captured from camera to solve tasks more efficiently \cite{munk2016learning}. Since SRL is usually conducted in an unsupervised manner, it demands advanced priors of the tasks.

Two mainstream works prove to address SRL effectively, in which we comprehensively review in our former survey\cite{lesort2018state}.  Contrastive learning methods \cite{chen2020simple, he2020momentum, bachman2019learning, hjelm2018learning} leverage the insight that similar states are closer to each other while dissimilar ones are pushed farther apart, to learn a lower-dimensional representation in an unsupervised manner. On the other hand, predictive methods leverage latent-space dynamics modeling as an auxiliary task. \cite{raffin2019decoupling, lange2012autonomous, finn2015learning} employ an auto-encoder to find a reduced set of features that encode necessary information to reconstruct the observations. Let $\phi$ be an encoder that encodes an observation $o_t$ to a latent state $s_t$, and $\psi$ a decoder to reconstruct the original observation from $s_t$. With $\theta$ being parameters of the functions, the optimization of encoder and decoder can be formulated as:
\[
\begin{split}
    s_t = \phi(o_t; \theta_\phi)\\
    \hat{o}_t = \psi(s_t; \theta_\psi)
\end{split}
\]
In the training process, by optimizing the loss function (\ref{equation_1}) iteratively, we can derive a state estimator that encodes essential information of input observations:
\begin{equation}
    \mathcal{L}_{reconstruct} = \frac{1}{2}(\lVert \hat{o}_t -o_t\rVert^2_2 + \lVert \hat{o}_{t+1} -o_{t+1}\rVert^2_2) \label{equation_1}
\end{equation}
\cite{gelada2019deepmdp} proposed to learn the environment dynamics via predicting future states coupled with reward prediction. However, such system dynamic models are highly non-linear and stochastic that often lead the learned latent space to collapse \cite{schwarzer2020data}. \cite{guo2020bootstrap} employs a pair of networks for both forward and inverse prediction. These works learn the observation-to-state mapping as part of learning some other functions such as the transition of the environment  $\hat{s}_{t+1} = f(s_t, a_t; \theta_{fwd})$, where $s_t = \phi(o_t)$. Thus, by minimizing $\mathcal{L}_{forward} = \lVert \hat{s}_{t+1} -s_{t+1}\rVert^2_2$ we can learn the encoder $\phi$ together with the transitional function $f$. This approach is addressed as the \emph{forward model}, while \emph{inverse model} and \emph{reward model} \cite{lesort2018state} are similar in concept but predict respectively the action $a_t$ that agent takes at $t$ and reward $r_{t+1}$ for next state using $s_t$ and $s_{t+1}$.
Similarly, \cite{schwarzer2020data} proposed a multi-step latent prediction task that enforce the long-term learned state representation. In robotics, to learn such state representation is rather intuitive, considering the physical priors we can benefit from. For example, robot operations are all deployed in real physical platform that could be considered as 2 or 3 dimension, while their intrinsic features are usually position, orientation, speed or acceleration. Authors in \cite{jonschkowski2014state} thus introduce some \emph{robotic priors} that only pertain to tasks involving robots in real-world function. Since manually designing priors are tedious, \cite{he2022reinforcement} proposed to automatically search for top-performing auxiliary loss functions for SRL.\par

In our former work \cite{raffin2019decoupling}, we notice that the performance can be significantly boosted by combining the above prediction sub-tasks altogether.


\subsection{Combining SRL Sub-tasks}

\begin{figure}[ht!]
    \centering
    \includegraphics[width = 0.48\textwidth]{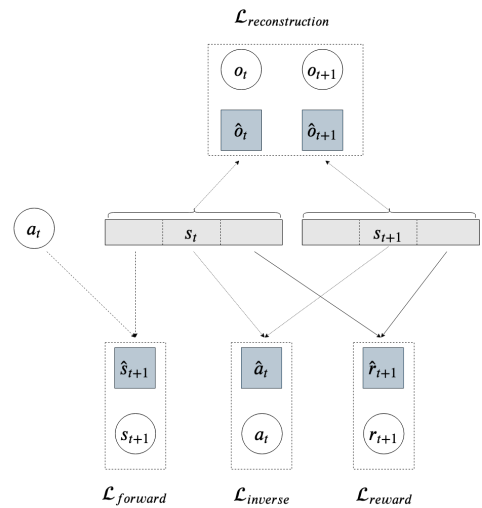}
    \caption{SRL: The whole state is used to reconstruct the observations but state space is split into three part for forward, inverse and reward model.}
    \label{fig:state_representation}
\end{figure}

In this work, we study how to effectively combine the prediction priors to effectively learn informative representation. We can either use the whole state to apply these different priors, or split the states and apply objective functions to different parts of the state space. One motivation to apply the \emph{srl split} model is that losses have different magnitudes and each prior have specialized functionality, optimization of all models on the same space often leads to collapse, leaving some features unlearned\cite{raffin2019decoupling}. Fig. \ref{fig:state_representation} demonstrates an example of SRL with split dimension. We use the whole state to reconstruct the observations yet split state space into three parts for forward, inverse and reward model. This can enforce components to specialize in encoding different featuring representations. \par

\par

By conducting SRL with above models, we can extract informative features into a lower dimension space. This process contributes on the one hand to accelerate the learning process so that it raises the possibility to train the RL agent directly on the robot. On the other hand, this low dimensionality offers an approach to interpret model extracted features, which facilitates the interpretability of deep RL methods.

\section{Policy Optimization via Online Abstract Representation}

Although SRL is confirmed to succeed in extracting some important features from high-dimensional sensory data, the state estimator derived from decoupling strategy usually deteriorates over the course of training\cite{munk2016learning, raffin2019decoupling}. This is because the SRL model is only pre-trained with randomly sampled frames, whose parameters stay the same throughout the rest of the training, leading to over-fitting. Therefore, this work is inspired to combine the previous SOTA SRL sub-learning tasks and integrate these priors together with \emph{domain resemblance prior} into the original PPO model, so as to update SRL adaptively throughout the training process. The design of POAR framework, \emph{domain resemblance prior} is introduced in section \ref{Sec:Framework}, section \ref{Sec:Loss_Function}. The general design of loss function and update strategy is explained in section \ref{Sec:General_design}.

\subsection{POAR Optimization Framework} 
\label{Sec:Framework}
On the basis of the original framework of PPO, we integrate the SRL model and introduce a dynamic mechanism of loss weighting (discussed in section \ref{Sec:General_design}) to guarantee stability and convergence. We incorporate several effective priors in the model network, which takes advantage of the observation-action tuple $(o_t, a_t)$ that PPO agent collects each batch to calculate loss based on predictions.\par

The proposed end-to-end state representation learning and reinforcement learning algorithm POAR is illustrated in Fig. \ref{fig:poar_structure}. The detailed architecture and hyperparameters can be found in Appendix \ref{tab:hyperparameter_poar}.

\begin{figure}[ht!]
    \centering
    \includegraphics[width = 0.65\textwidth]{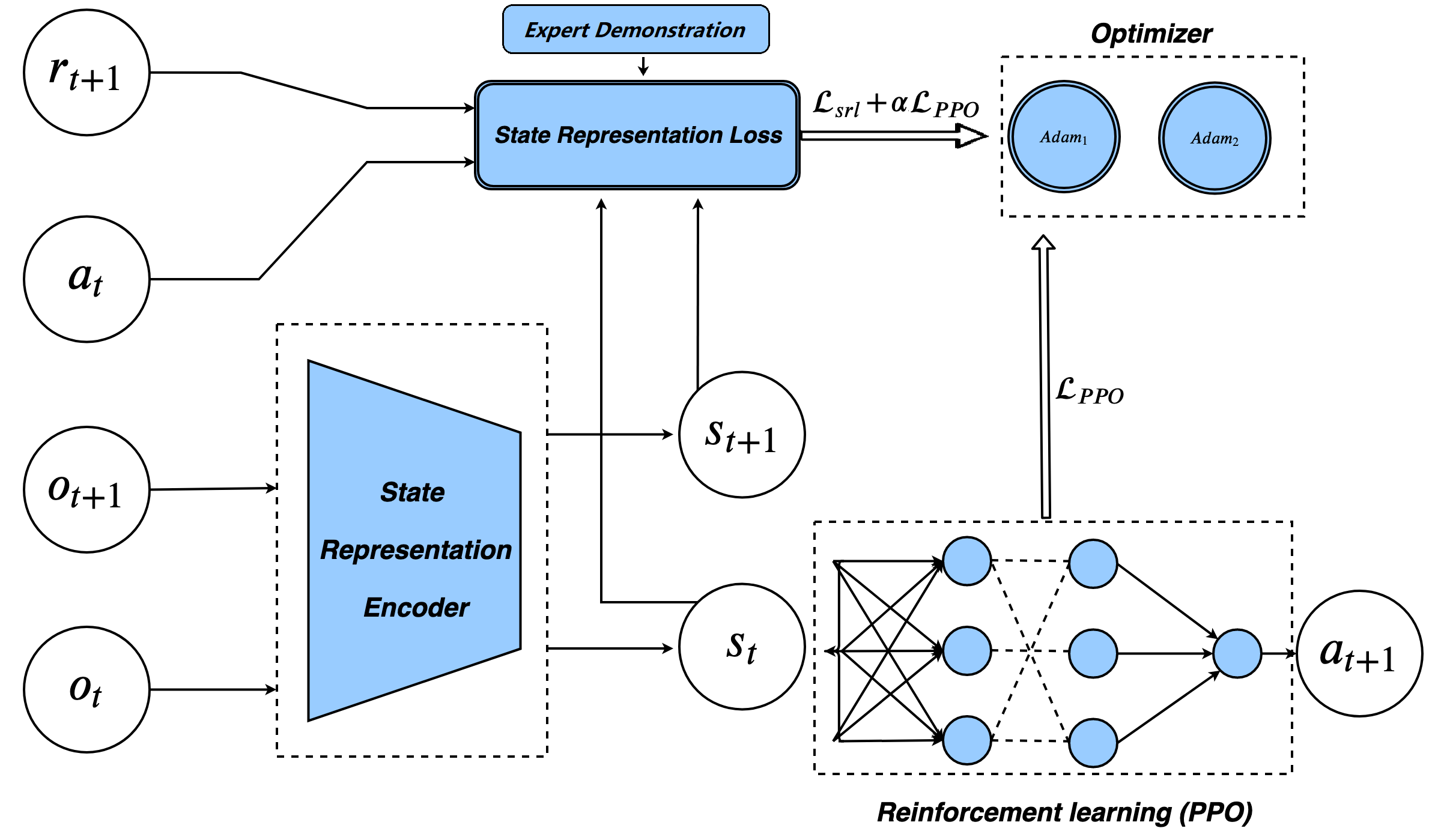}
    \caption{The design of our proposed framework POAR: Policy Optimization with Online Abstract State Representation Learning, an end-to-end SRL-RL training pipeline.}
    \label{fig:poar_structure}
\end{figure}

In order to preserve the learned states to be representative for the physical meaning, we use two \emph{Adam} optimizers (adaptive moment estimation) \cite{kingma2014adam} that update separately the RL and SRL models. We add $\mathcal{L}_{SRL}$ to assist in RL loss so that the states can evolve to meet the demand of reinforcement learning and maintain a good physical interpretability. By engaging RL loss to update SRL model, which pertains to learning the encoder $\phi$ together with another function $\pi^*$, the state estimator is thus more robust and task-relevant.\par
This framework guarantees the timely update of SRL model, by which we obtain an effective and active representation of states. With the help of \emph{domain resemblance prior}, our model can further boost the sample efficiency of RL training.

\subsection{Domain Resemblance Prior}
\label{Sec:Loss_Function}
Simulation results show that directly using the task-relevant coordinates of robots as states to train RL agent can generally obtain better result than training from raw images\cite{xie2023simulation, kober2013reinforcement}. Therefore, we conceive the coordinate itself as a representational state, and seek to learn a mapping from raw image to ground-truth coordinates. To further guide the policy, we leverage expert demonstration as prior knowledge\cite{chen2021efficiently}. Thus we characterize portions of SRL states with robots' coordinates in real-world (2D or 3D) and enforce the learned states resemble the distribution of expert trajectories. Considering that it's not practical to provide accurate coordinate states for each camera frame when training a real-life robot, we apply \emph{Maximum Mean Discrepancy} (MMD) \cite{gretton2012kernel} to measure the distribution deviation between sampled exploration and demonstration trajectories:
\begin{equation}
    \mathcal{L}_{dr}= \vert\vert\frac{1}{n}\sum_{i=1}^{n}\Phi(\mathcal{D}_{s})-\frac{1}{m}\sum_{j=1}^{m}\Phi(\mathcal{D}_{d})\vert\vert_{\mathscr{H}}^{2}
    \label{MMD}
\end{equation}
where $\Phi(\cdot)$ maps the original data to the regenerated \emph{Hilbert} space \cite{berlinet2011reproducing}. By back-propagating the MMD loss, the learned states can effectively restore distribution of coordinates in the real work space, and also guide the RL agent to imitate expert's behaviors\cite{finn2016guided}. We mainly use this prior to aid \emph{forward} model on the same state portion to build a more structural understanding of the environment space. In conclusion, domain resemblance prior can facilitate the reconstruction of observations and increase interpretability of the learned state estimator.

\subsection{Design of Loss Function and Update Strategy}
\label{Sec:General_design}
With the presence of \emph{Domain Resemblance} prior and some other useful priors like \emph{forward}, \emph{inverse} and \emph{autoencoder}, the SRL model is promising to learn an efficient and task-relevant representation with strong physical interpretations. The loss function can be formulated as:
\[
\begin{split}
\mathcal{L}_{RL} & =\hspace{7.3ex} \mathcal{L}_{PPO}  \hspace{6ex}\hspace{14ex} (Optimizer 1)
\\
\mathcal{L}_{SRL}& = \sum_{x\in\mathrm{SRL\, Model}} w_{x} \times \mathcal{L}_{x} \hspace{13.6ex}(Optimizer 2) \\
& = w_{iv}\mathcal{L}_{iv} + w_{fw}\mathcal{L}_{fw} + w_{rw}\mathcal{L}_{rw} + w_{ae}\mathcal{L}_{ae} + w_{dr}\mathcal{L}_{dr}
\end{split}
\]
where $w_{x}$ denotes the weights attributed to each SRL sub-learning tasks.\par
Nevertheless, a direct optimization on two losses is prone to deteriorate the encoder. The overall model parameters can be back-propagated by both the SRL encoder and RL losses, while the latter are generally larger than that of SRL, which might lead to the collapse (reduced to a point or strongly deformed) of states\cite{lesort2018state}. Therefore, we reduce scales of RL gradients on encoder part for \emph{Optimizer 1}. Let denote $\theta$ as parameters, $\Theta_{SRL}$, $\Theta_{RL}$ as the sets of parameters respectively concerning SRL model and RL model. If $\theta \in \Theta_{SRL}\cap \Theta_{RL}$, then:
\begin{equation}
    Grad_{\mathrm{Optimizer}_1} (\theta) = \alpha \times \frac{\partial\mathcal{L}_{RL}}{\partial\theta}
\end{equation}
where $\alpha$ is a hyperparameter to balance the trade-off on optimization of PPO and SRL model on encoder part. Later in section \ref{Sec:Ablation_Study} we present an ablation study on the best choices of $\alpha$. We notice in our experiments that the SRL model generally converges long before the RL policy (the randomness of the policy at the early stage of RL contributes to the fully optimization of SRL models while exploring the environment). Once the SRL model is optimized, we seek to maintain its properties by applying a different learning rate decay strategy for two optimizers: the learning rate of PPO model  $lr_1$ decays linearly to zero till the end while the learning rate of SRL model $lr_2$ exhibits an exponential decay slope that decreases much faster.
\begin{equation}
lr_{1} = lr^1_0 \times r; \quad lr_2 = lr^2_0 \times \max(\exp{(-\beta r)},\hspace{0.5ex} 0.001 r); 
\end{equation}
where $r = 1- \frac{n-1}{N}$, $n$ is the current time steps and $N$ denotes the total number of time steps. $\beta$ denotes another hyperparameter to control the weight decay of SRL model's learning rate. We elaborate in section \ref{sec:exp} on the effectiveness of this dynamic weighting strategy on guaranteeing the convergence and optimality of the learned policy.
\section{Experimental Results}
\label{sec:exp}
\begin{figure}[ht!]
  \centering
    \subfigure[\emph{MobileRobot}]{\includegraphics[width=0.225\textwidth]{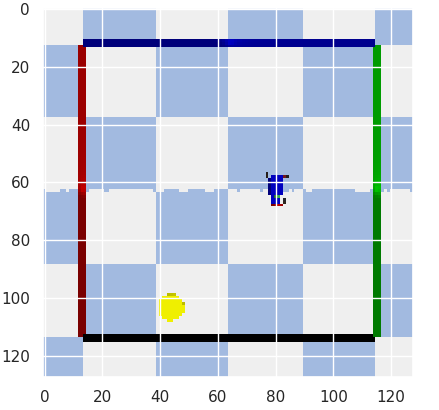}} 
	\subfigure[\emph{Omnirobot}]{\includegraphics[width=0.225\textwidth]{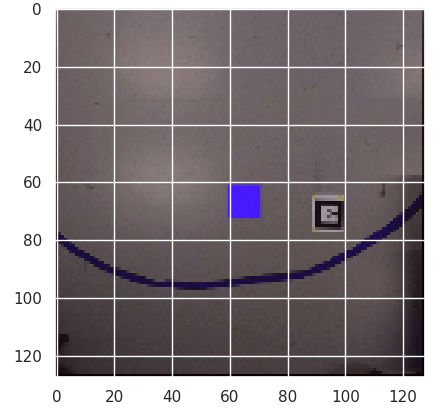}}
	\subfigure[\emph{Jaka}]{\includegraphics[width=0.21\textwidth]{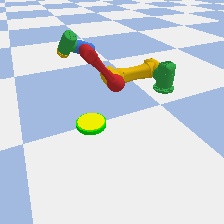}}
	\subfigure[\emph{In Real-life}]{\includegraphics[width=0.235\textwidth]{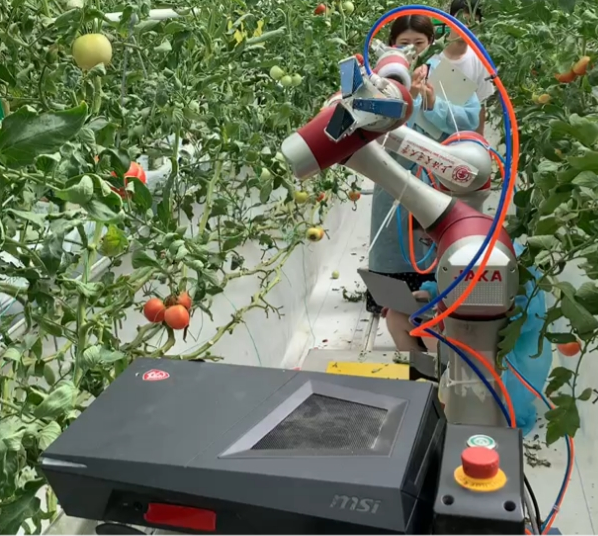}}
	\caption{Robotics environments that we select to perform experiments. \emph{MobileRobot} aims to reach the yellow target with as few steps as possible. \emph{Omnirobot} learns to circle the target by optimizing (\ref{equation:cc_reward}). \emph{Jaka} is a 6-DOF mechanical robot manipulator.}
	\label{fig:Environments_Demonstration}
\end{figure}

To comprehensively assess the performance of POAR, we select 3  robotic environments: \emph{MobileRobot} which consists of a racing car navigating to reach a target in 2D, \emph{Omnirobot} which consists of a robot circling around the target in Mujoco\cite{todorov2012mujoco} and a mechanical robot manipulator \emph{Jaka}, as illustrated in Fig. \ref{fig:Environments_Demonstration}. For the ease of operation, we perform experiments on a simulation platform \emph{RobotDrlSim} \cite{sun2021robotdrlsim} developed by our lab. Please check the github repository \footnote{\url{https://github.com/BillChan226/POAR-SRL-4-Robot}} for the algorithm implementation and the simulator. The comparison results are 3-folded: (1) POAR is first compared to SOTA deep RL algorithms \cite{schulman2017proximal} to assess the necessity of an explicit SRL feature extraction model; (2) POAR is compared to decoupling SRL methods\cite{raffin2019decoupling} to demonstrate the effectiveness of evolving SRL models online to adapt to RL policy; (3) We also compare the performance of splitting and combining SRL priors for a stationary state space.

\subsection{Quantitative Analysis}
The robot in \emph{MobileRobot} aims to reach the yellow target with as few steps as possible, by which it receives a positive reward (+1), while receiving a negative penalty (-1) when it hits the boundaries. Based on three different random seeds, we compare POAR model with both split and combination optimization strategy to end-to-end PPO baseline and decoupling SRL with split strategy presented in \cite{raffin2019decoupling}. 

\begin{figure}[ht!]
  \centering
    \subfigure[Performance comparison in \emph{MobileRobot}.] {\includegraphics[width=0.45\textwidth]{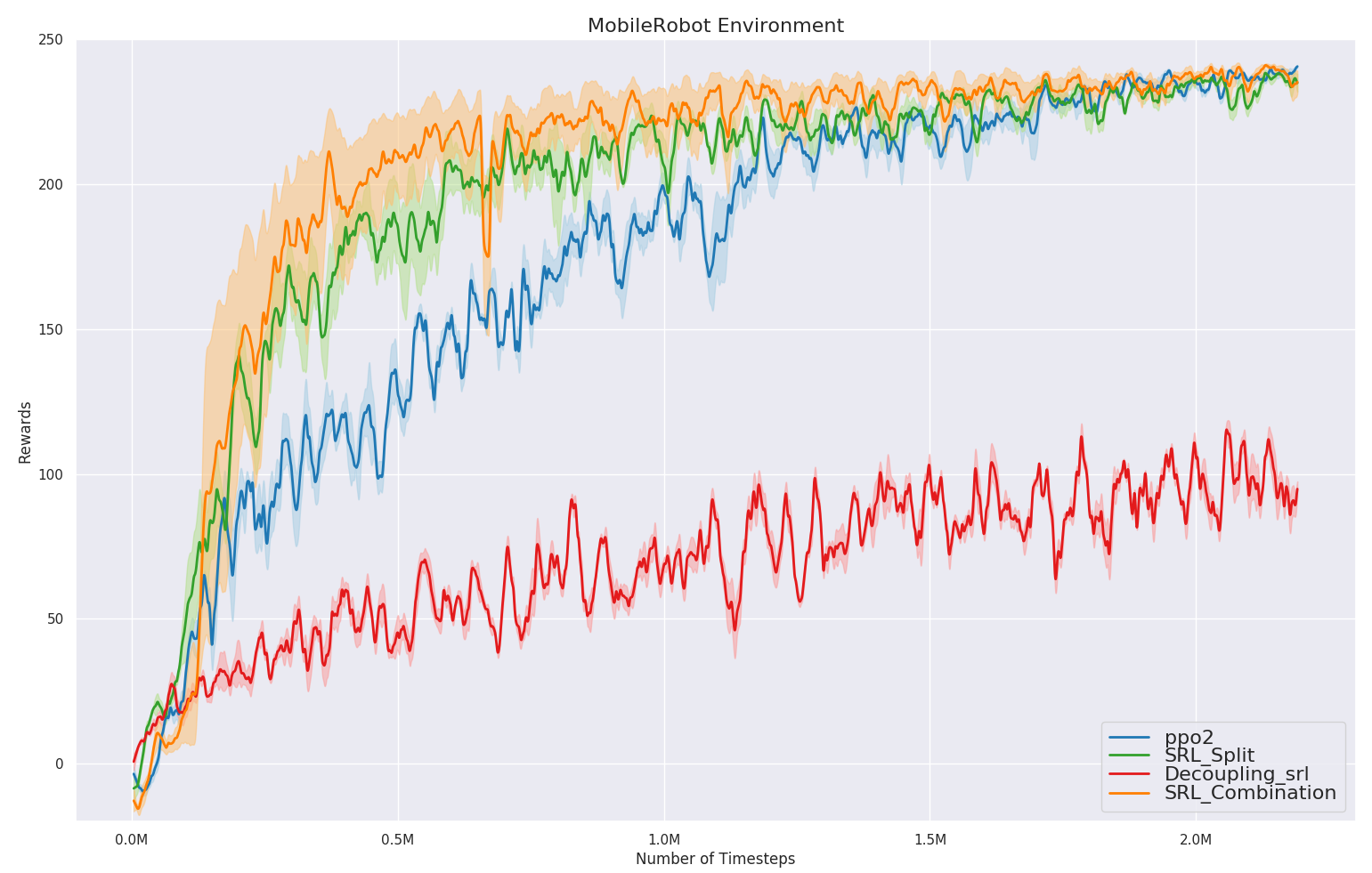}\label{Com1}} 
	\subfigure[Ablation study in  \emph{Omnirobot}.]{\includegraphics[width=0.43\textwidth]{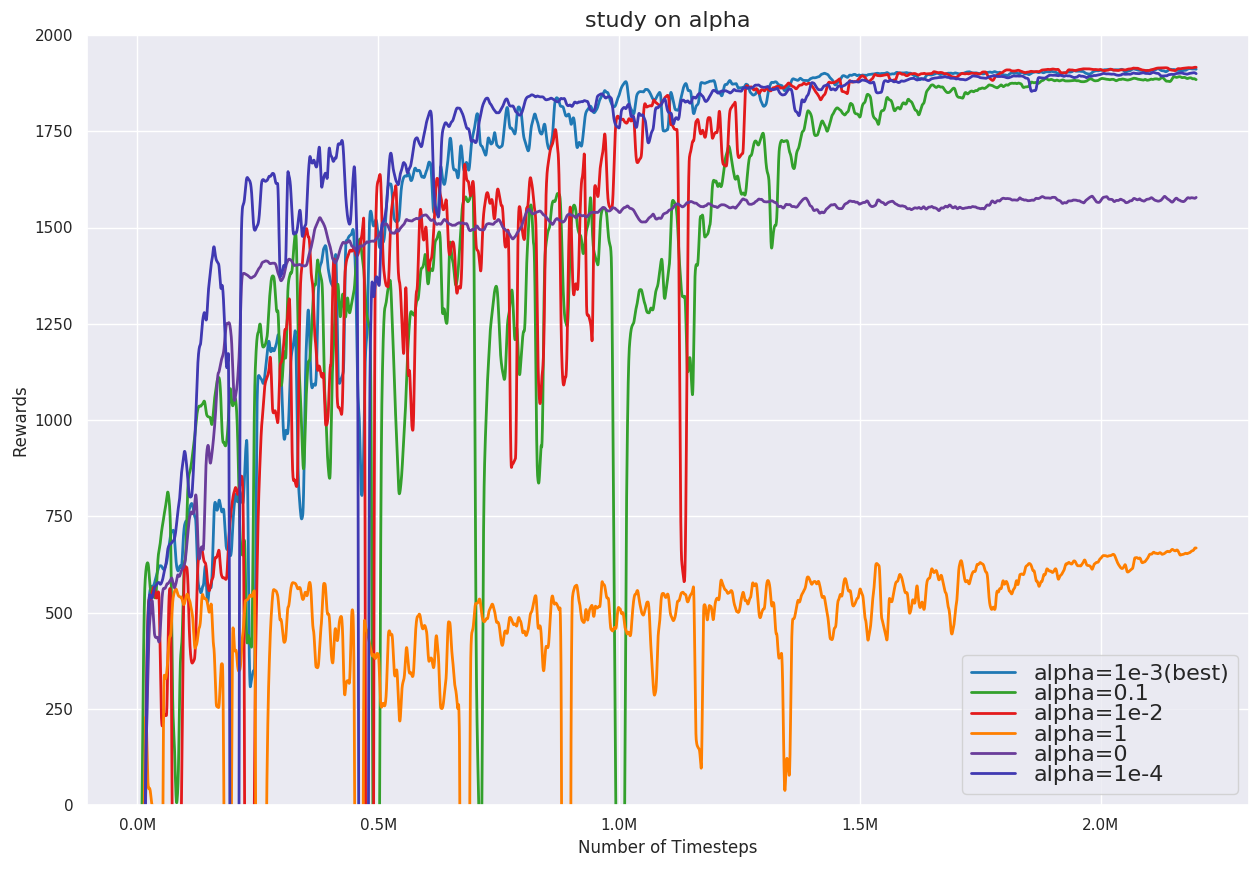}\label{Com2}} \\
	\caption{\subref{Com1}: Results which compare POAR with original PPO and decoupling SRL method. \subref{Com2}: A study on trade off parameter $\alpha$. }
	\label{fig:MobileRobotStudy}
\end{figure}

The improvement in performance of POAR model regardless of optimization strategy is illustrated in Fig. \ref{Com1}. The decoupling strategy is sensitive to splitting dimension due to the lack of a good metric to evaluate the SRL model, leading to unstable performance and over-fitting. On the contrary, since POAR can update to adapt to the task while RL agent explores to learn, it exhibits better convergent accuracy.\par

\emph{Omnirobot} learns to circle the target by optimizing (\ref{equation:cc_reward}). 
\begin{equation}
    r_t = \lambda \times (1-\lambda(\lVert z_t\rVert-R)^2) \times \lVert z_t - z_{t-k}\rVert^2_2 + \lambda^2 r_{t, bumped}
    \label{equation:cc_reward}
\end{equation}\par
A main difference that distinguishes \textit{Circling} from \textit{Reaching} task is that the robot is encouraged to move away from the position after $k$ steps. In \emph{Omnirobot}, the robot learns not only how to circle the target but also how to avoid bumping into walls, namely to learn to follow a specific trajectory, which introduces more dynamical features. We carried out a set of thorough experiments to compare the SOTA RL algorithms on this typical environment. 

\begin{figure}[ht!]
    \centering
    \includegraphics[width = 0.65\textwidth]{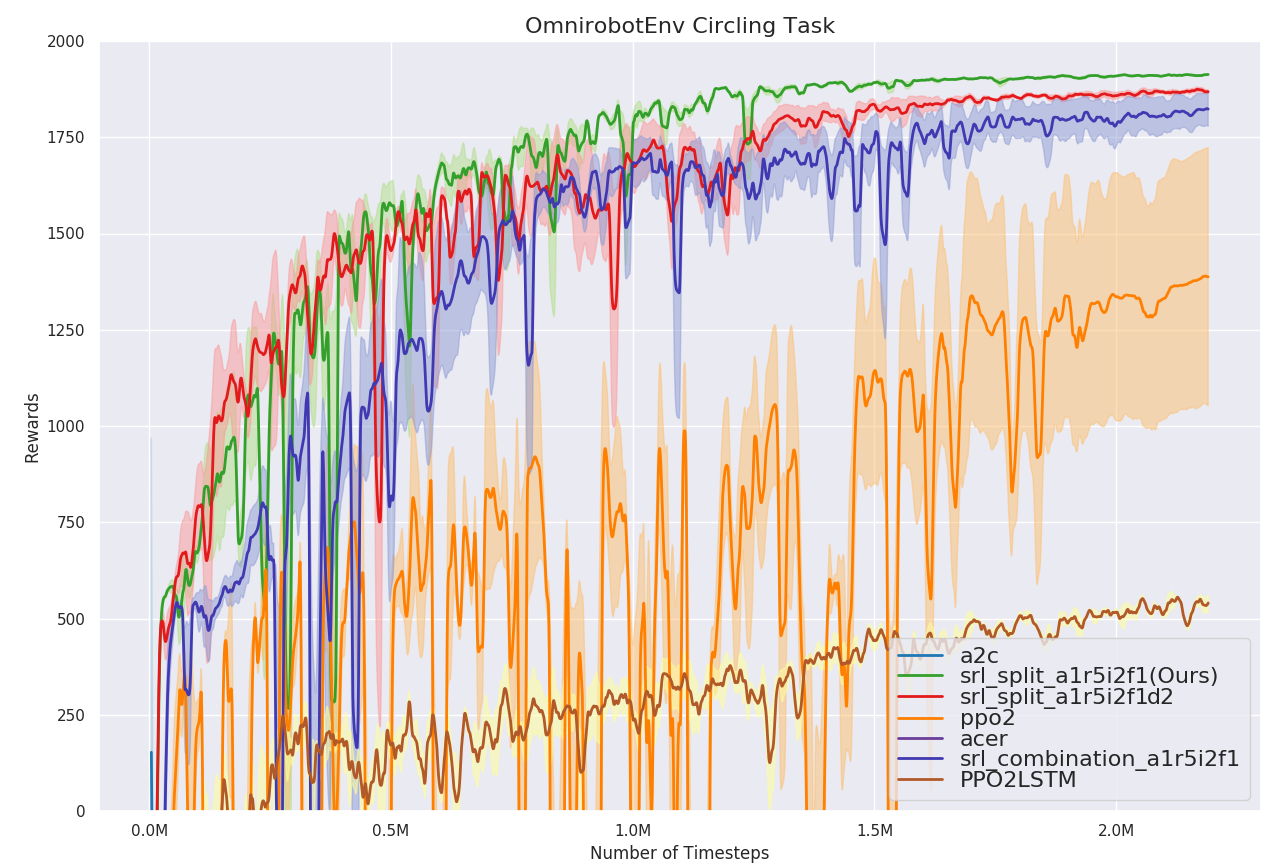}
    \caption{Results compared with PPO2, ACER and A2C in environment \textit{Omnirobot}, 2D navigation task where the robot aims to circle around the blue square. Results are based on three different random seeds. Bright color domain denotes one standard deviation around mean value.}
    \label{fig:poar_Omnirobot_results}
\end{figure}

The results are depicted in Fig. \ref{fig:poar_Omnirobot_results}. The legends of curve denote the weights assigned to different state representation models. For instance, \emph{a10r5f1d2} denotes $w_{autoencoder}, w_{reward}, w_{forward}, w_{domain} = 10, 5, 1, 2$. 
Table \ref{tab:policy_regrets} is a quantitative analysis of Fig. \ref{fig:poar_Omnirobot_results}. 

\begin{table}[ht!]
\scriptsize
\centering
\caption{Quantitative Analysis on sample efficiency and rewards}
\begin{tabular}{lcc}
\textbf{SRL model} & \textbf{Policy Regrets ($\pm$ std)} & \textbf{Rewards ($\pm$ std)}\\ \hline
PPO &  1.000 ($\pm$ 0.096) & 1333.638 ($\pm$ 565.189)\\ \hline
PPO with LSTM policy   & 0.936 ($\pm$ 0.025) & 497.529 ($\pm$ 24.993)\\ \hline
ACER   & 14.576 ($\pm$ 0.003) & -24220.244 ($\pm$ 73.846)\\ \hline
A2C  & 14.506 ($\pm$ 0.015) & -24306.119 ($\pm$ 73.744)\\ \hline
SRL-comb decoupling   & 0.078 ($\pm$ 0.019) & 1855.563 ($\pm$ 24.153)\\ \hline
SRL-split decoupling    & 0.080 ($\pm$ 0.009) & 1845.733 ($\pm$ 46.225)\\ \hline 
\textbf{SRL Split(Our)}   & \textbf{0.164} ($\pm$ \textbf{0.028}) & \textbf{1909.470} ($\pm$ \textbf{3.371})\\ \hline 
SRL Combination   & 0.293 ($\pm$ 0.059) & 1787.086 ($\pm$ 101.075)\\ \hline
\end{tabular}
\label{tab:policy_regrets}
\end{table}

We can denote that POAR outperforms SOTA RL algorithms in terms of sample efficiency (evaluated with \emph{Policy Regret}\footnote{Policy regret, defined as the difference between the cumulative reward of the optimal policy and that gathered by $\pi$. $\int$\emph{target reward} $- \mathbb{E}[r^{\pi}(s)] \mathrm{d}s$. A smaller policy regret indicates a better sample efficiency.}), stability and final rewards. By comparing the variation and smoothness of the curves, we conclude that POAR improves stability and enhance robustness of convergence. On the contrary, PPO with LSTM policy \cite{graves2012long} fail to exhibit an excellent performance in this sequential task as expected (due to the curse of dimensionality\cite{bellman1966dynamic}). Contrarily, thanks to the fact that POAR is equipped with forward, inverse, reward and domain resemblance priors, it can thus efficiently model the environment space and capture the dynamical changes within time steps.

\begin{figure}[ht!]
  \centering
    \subfigure[\label{Com3}]{\includegraphics[width=0.45\textwidth]{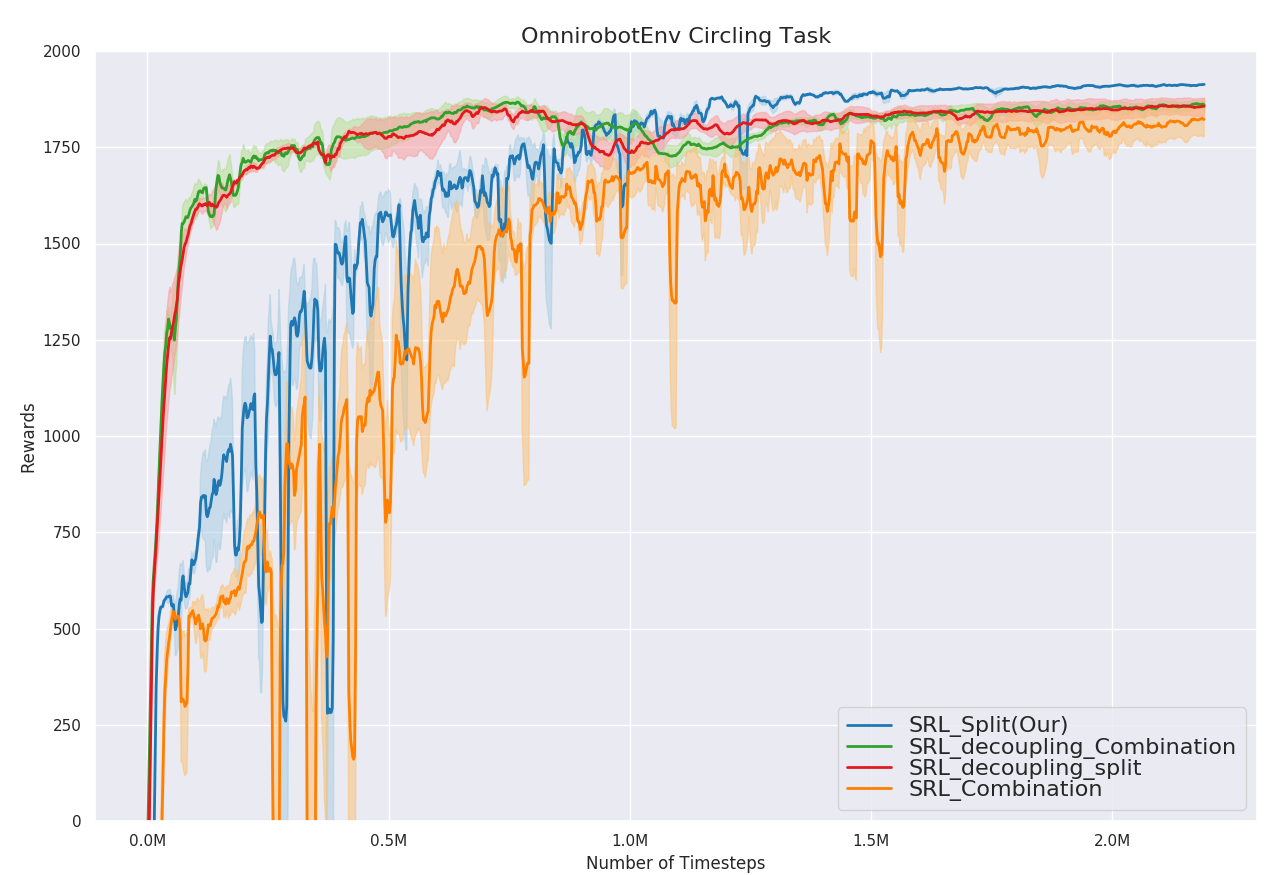}} 
	\subfigure[\label{Jaka}]{\includegraphics[width=0.42\textwidth]{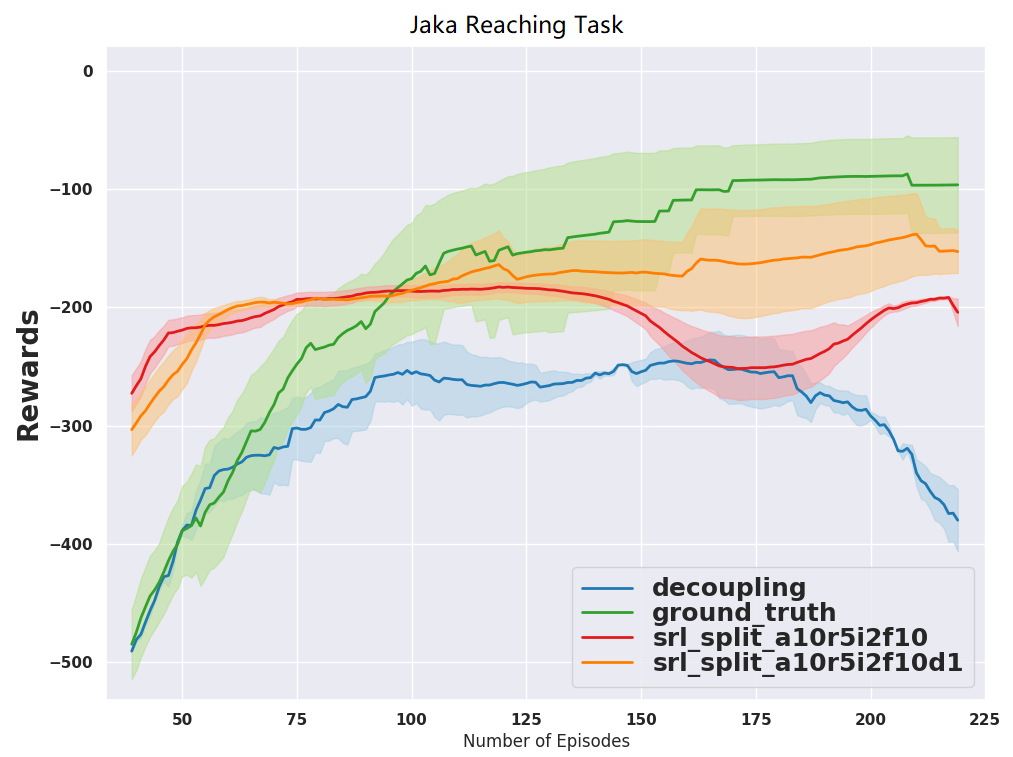}} \\
	\caption{\subref{Com3}: results in \emph{Omnirobot} which compare POAR with decoupling SRL strategy of split and combination weights attribution. \subref{Jaka}: results in \emph{Jaka} that demonstrates effectiveness of domain resemblance prior.}
	\label{fig:Decoupling_Compare_Study}
\end{figure}

Fig. \ref{Com3} compares POAR to the decoupling strategy with both split and combination mode. During early training stages, decoupling strategies demonstrate a strong sample efficiency regardless of split or combination mode. However, they suffer from over-fitting over the state representation at the final stage of RL training, due to the lack of exploration. In addition, the abundant pre-sampling guarantees the convergence rate at the beginning of training for decoupling strategies, otherwise the bias as illustrated in Fig. \ref{Decoupling_lack} is difficult to overcome. Thus generally, POAR exhibits better sample efficiency, since the sampled data is leveraged to train both SRL and RL models.\par

We also compare performance on a more complex and dynamic environment \emph{Jaka}, in which the robot tries to push the button on the ground. Results in Fig. \ref{Jaka} show that ground-truth coordinate is an informative enough state representation from which we can train an efficient agent. Thus it explains why the model with \emph{domain resemblance} can outperform the one without, since a more structural perception of the environment space is built into the states.

\subsection{Qualitative Analysis}
\begin{figure}[ht!]
  \centering
    \subfigure[Original image to encode]{\includegraphics[width=0.3\textwidth]{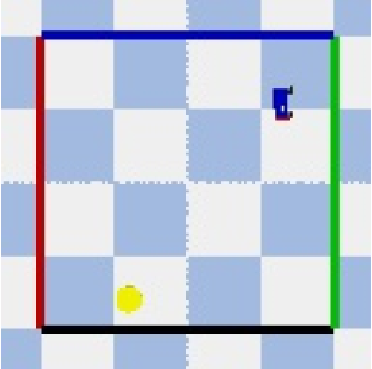}} 
	\subfigure[Reconstructed image]{\includegraphics[width=0.3\textwidth]{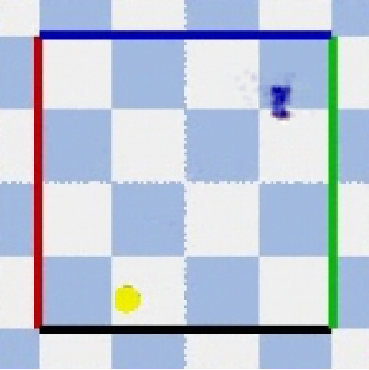}} \\
	\caption{Reconstruction result in \emph{MobileRobot} by introducing domain resemblance prior.}
	\label{fig:Reconstruction}
\end{figure}

In this subsection we provide an access for human-in-the-loop inspection, namely by monitoring the state graph iteration to diagnose current policy and SRL model. Firstly we show in Fig. \ref{fig:Reconstruction} the reconstruction result of \emph{MobileRobot} by introducing \emph{domain resemblance} loss.

\begin{figure}[ht!]
  \centering
    \subfigure[POAR model]{\label{fig:2D}\includegraphics[width=0.3\textwidth]{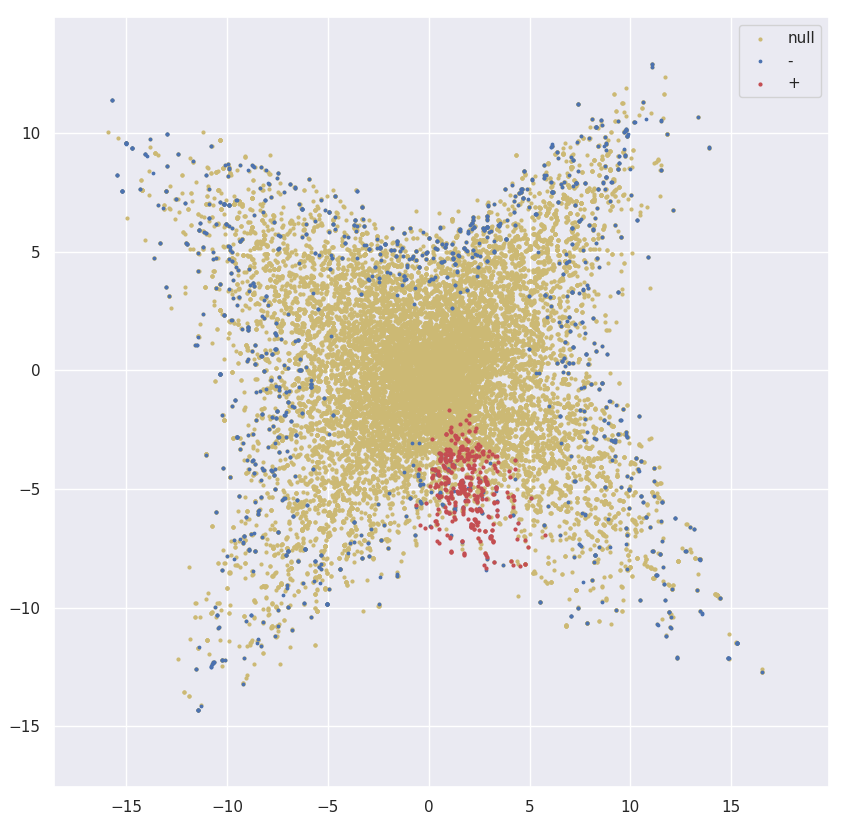}}
	\subfigure[Decoupling strategy]{\label{Decoupling_lack}\includegraphics[width=0.42\textwidth]{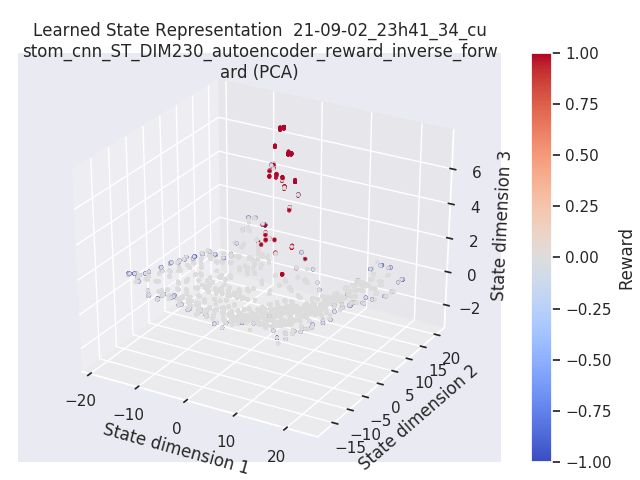}} \\
	\caption{Comparison between POAR and decoupling stategy of learned states in \emph{MobileRobot}, respectively cast into 2D and 3D.}
	\label{fig:Learned_States}
\end{figure}

With such priors engaged, the interpretation of latent states is further enforced. Fig. \ref{fig:2D} is a 2D projection, using PCA, of the state inferred by the SRL encoder, which we obtained via a policy at the beginning of training POAR (at episode 200). To better demonstrate and compare the rewards' distribution, we project the learned states obtained by decoupling strategy to a 3D space, as shown in Fig. \ref{Decoupling_lack}.

\begin{figure}[ht!]
    \centering
    \begin{minipage}{0.3\textwidth}
    \centering
    \includegraphics[width=\textwidth, height=1\textwidth]{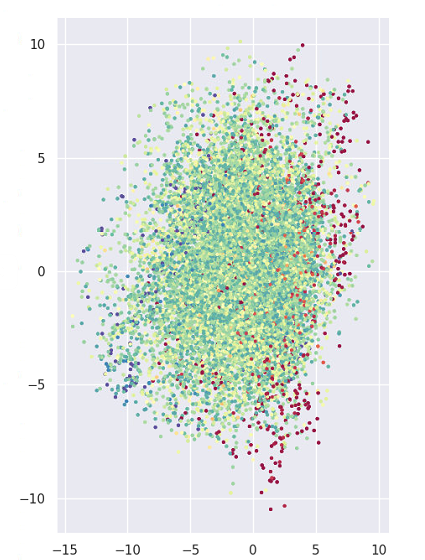}
    Episode 50
    \end{minipage}
    \begin{minipage}{0.3\textwidth}
    \centering
    \includegraphics[width=\textwidth, height=1\textwidth]{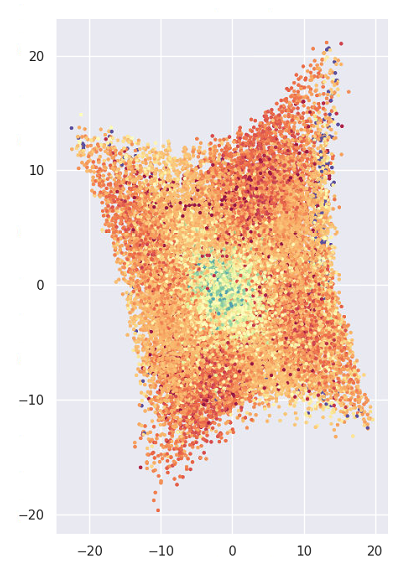}
    Episode 1000
    \end{minipage}
    \begin{minipage}{0.3\textwidth}
    \centering
    \includegraphics[width=\textwidth, height=1\textwidth]{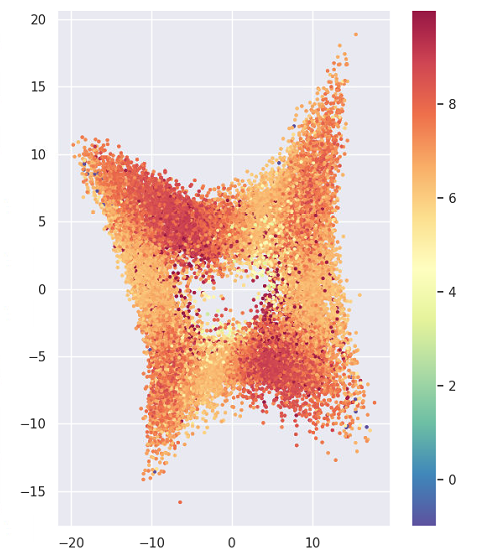}
    Episode 3000
    \end{minipage}
    \caption{The evolution of state graph in \emph{Omnirobot} through the course of RL training with POAR model. Each episode consists of 250 time steps of interactions. We used split dimension with $dim_{reward}: 120, \, dim_{inverse}:50, \, dim_{forward}:50$. The weights assigned on such SRL model are: $w_{autoencoder}:1,\,w_{reward}:1,\, w_{inverse}:5, \, w_{forward}:5$.}
    \label{fig:circular_state}
\end{figure}

States learned with decoupling strategy (Fig. \ref{Decoupling_lack}) is sparser and have some holes at which the state estimator can't offer the agent with sufficient information. 
Similarly, Fig. \ref{fig:circular_state} presents the evolution of state graph with POAR in \emph{Omnirobot}.

At $1000^{th}$ episode, we can observe the square boundary shape and reward distribution. At $3000^{th}$ episode, we reach a policy that achieves reasonable rewards (1830 scores within an episode). From the state at last checkpoint, we have a better understanding of the agent’s behavior: there is a specific direction that accumulates more rewards at four corners, which incites agent to circle around the center. 

\subsection{Ablation Study}
\label{Sec:Ablation_Study}
As is shown in Fig. \ref{Com2}, we study how $\alpha$, a hyperparameter that balances the trade off between RL and SRL loss, influences the training result. When $\alpha = 0.001$, the SRL model can best leverage the feedback from RL to improve itself, speeding up sample efficiency and improving stability.

Table \ref{tab:policy_regrets_mobile} shows how the performance of POAR varies with respect to different weights attribution. We can denote that almost all weight choices outperform the PPO baseline, in terms of sample efficiency and final rewards.

\begin{table}[ht!]
\scriptsize
\centering
\caption{Policy regret normalized by PPO and rewards in \textit{MobileRobot}.} 
\begin{tabular}{cccccc}
$\bm{w}_{a}$ & $\bm{w}_{r}$ & $\bm{w}_{i}$ & $\bm{w}_{f}$ & \textbf{Policy Regrets}& \textbf{Rewards ($\pm$ std)}\tnote{1} \\ \hline
1 & 0 & 0 &0   & 1.128 ($\pm$ 0.065) \\ \hline 
1&0&1&5  & 0.880 ($\pm$ 0.019) & 237.94 ($\pm$ 4.06)\\ \hline
10&0&1&1    & 0.869 ($\pm$ 0.140) & 240.32 ($\pm$ 1.11)\\ \hline
1&0&10&1   & 0.907 ($\pm$ 0.049) & 239.99 ($\pm$ 2.69)\\ \hline
1&0&1&1    & 0.940 ($\pm$ 0.105) & 234.75 ($\pm$ 3.22)\\ \hline
1&0&1&10    & 1.012 ($\pm$ 0.076) & 237.20 ($\pm$ 0.65)\\ \hline
5&0&1&1   & 0.843 ($\pm$ 0.073) & 232.12 ($\pm$ 6.12)\\ \hline
1&0&5&1    & 0.814 ($\pm$ 0.067) & 239.75 ($\pm$ 3.18)\\ \hline
5&0&1&1   & 0.843 ($\pm$ 0.073) & 232.12 ($\pm$ 6.12)\\ \hline 
1&5&2&1   & 2.406 ($\pm$ 0.042) & 94.90 ($\pm$ 1.04) $^d$\\ \hline 
1&5&2&1   & \textbf{0.684} ($\pm$ \textbf{0.102}) & \textbf{236.21} ($\pm$ \textbf{3.19}) $^s$\\ \hline
1&5&2&1  & \textbf{0.562} ($\pm$ \textbf{0.228}) & \textbf{235.90} ($\pm$ \textbf{4.32}) $^c$\\ \hline 
\\ \hline
PPO & (Baseline) & &   & 1.000 ($\pm$ 0.063) & 234.242 ($\pm$ 3.867)\\ \hline 
\end{tabular}
\begin{tablenotes}
       \footnotesize
       \item[1] $d$ denotes decoupling strategy; $s$ denotes split mode; $c$ denotes combination mode.
\end{tablenotes}
\label{tab:policy_regrets_mobile}
\end{table}

\begin{figure}[ht!]
    \centering
    \includegraphics[width = 0.8\textwidth]{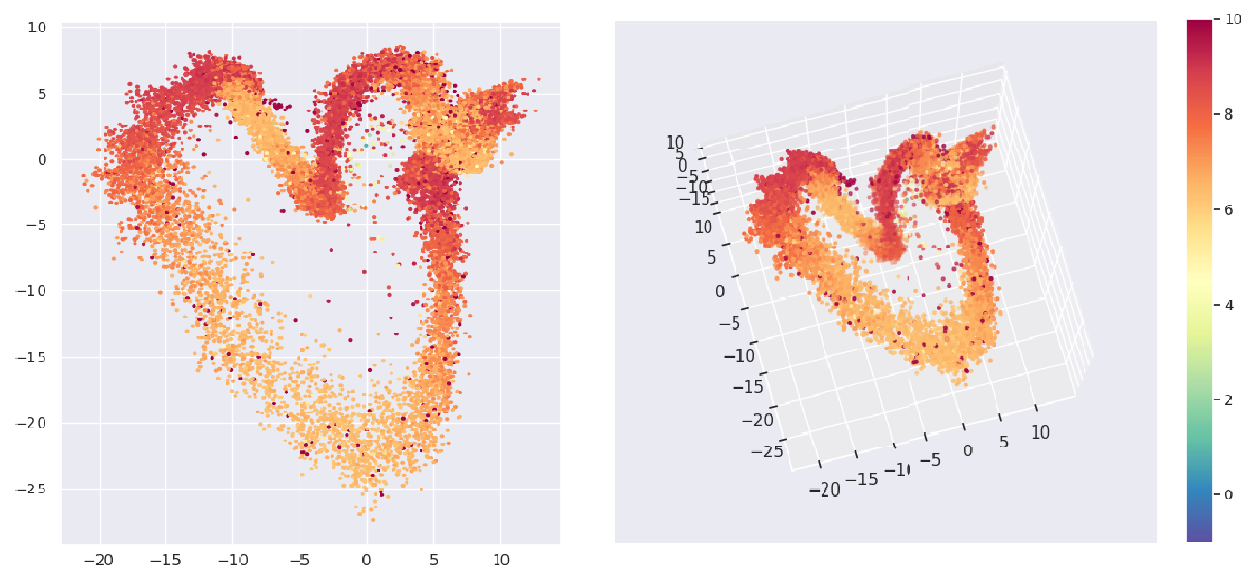}
    \caption{State graph with different SRL weight attribution in \emph{Omnirobot}.}
    \label{fig:States_Comparison}
\end{figure}

Fig. \ref{fig:States_Comparison} shows inferred abstract state trained with SRL weights partition: $w_{reward}:5, \, w_{inverse}:2, \, w_{forward}:1$ and with the same split dimension strategy as Fig. \ref{fig:circular_state}. The bias on reward model encourages state to separate points with different rewards in higher dimension. Although it has a less physical meaning for robotic, yet, it has a better sample efficiency during training.
\section{Conclusions}

To conclude, we summarize the SOTA SRL sub-tasks in prior works and present a new framework to train RL agent together with the SRL model, as well as a \emph{domain resemblance prior} to leverage expert demonstration. With such features incorporated, our framework can efficiently handle the tasks in high-dimensional robotic scenarios, and encourages the potential to train real-life robots from scratch. During training, POAR can provide real-time access to the current inferred states which helps to interpret RL with a real physical meaning, and we can thus monitor the course of training and diagnose the policy. In detail,

\begin{enumerate}
    \item Our model integrates SRL into the training process of RL which allows SRL to adapt with RL. Learned in this way, the observation-state mapping can thus guarantee the high final rewards of RL, since over-fitting is overcome.
    \item By reusing the interactions collected by RL for training SRL and leveraging expert demonstration to guide the agent, the cold-start problem is tackled and sample efficiency is substantially sped up.
    \item Compared with the decoupling strategy, our model is more robust, almost all SRL models outperform the original PPO model in our setups. Hence, it is easier to tune the hyperparameter.
\end{enumerate}

In terms of future work, we would like to thoroughly test our algorithm on other standard RL frameworks to explore the general performance of POAR. Inspired by this framework, we could also develop other RL architectures based on A3C, ACER and deepq algorithms to further boost the sample efficiency of reinforcement learning.
\backmatter



\begin{appendices}

\section{Hyperparameters}\label{secA1}
Hyperparameter settings please refer to Table \ref{tab:hyperparameter_poar}. All other hyperparameters that are not presented on the table are the same as the PPO2 implementation of \emph{baselines} \cite{dhariwal2017openai}.
\begin{table}[h]
\centering
\begin{tabular}{lc}
\textbf{Hyperparameter} & \textbf{Value} \\ \hline
lr (Optimizer 1) & 5e-4  \\ \hline
lr (Optimizer 2) & 2.0e-4  \\ \hline
$\alpha$ & 0.001 \\ \hline
$\beta$ & 20\\ \hline
\end{tabular}
\caption{POAR hyperparameter settings}
\label{tab:hyperparameter_poar}
\end{table}

\end{appendices}


\bibliography{sn-bibliography}

\section*{Declarations}

\subsection{Funding}
This work was supported by Shanghai Agriculture Applied Technology Development Program, China (Grant No.C2019-2-2) and National Key Research and Development Program“Robotic Systems for agriculture(RS-Agri)” (2019YFE0125200).

\subsection{Competing Interests}
The authors have no relevant financial or non-financial interests to disclose.

\subsection{Author Contributions}

\textbf{Zhaorun Chen}: Conceptualization, Investigation, Formal analysis, Writing (original draft). \textbf{Liang Gong, Binhao Chen}: Investigation, Formal analysis, Writing (original draft). \textbf{Siqi Fan, Yuan Tan}: Investigation, Formal analysis. \textbf{Te Sun, David Filliat, Natalia Díaz-Rodríguez}: Conceptualization, Writing (review \& editing), Supervision. \textbf{Chengliang Liu}: Project administration, Funding acquisition.

\subsection{Ethics approval}
Not Applicable.
\subsection{Consent to participate}
Not Applicable.
\subsection{Consent to publish}
Not Applicable.

\end{document}